\documentclass[letterpaper, 10 pt, conference]{ieeeconf}  

\IEEEoverridecommandlockouts                              

\usepackage{graphicx}
\usepackage{algorithm}
\usepackage{algpseudocode}
\usepackage{silence}
\usepackage{caption}
\usepackage{amsmath} 
\usepackage{amssymb}  

\usepackage{verbatim}

\renewcommand{\vec}{\boldsymbol}

\newcommand{\gripper}{\mathcal{G}}
\newcommand{\objects}{\mathcal{O}}
\newcommand{\object}{o}
\newcommand{\pregrasps}{\mathcal{P}_o}
\newcommand{\pregrasp}{\vec{p}_o}

\newcommand{\cts}{\mathbf{c}} 
\newcommand{\enc}{\mathbf{q}} 
\newcommand{\pose}{\mathbf{x}} 
\newcommand{\twist}{\dot{\pose}} 

\newcommand{\physicsscore}{P\ }
\newcommand{\constanttorque}{CT\ }
\newcommand{\withcont}{C\ }
\newcommand{\withoutcont}{$\neg$C\ }
\newcommand{\withnoise}{N\ }
\newcommand{\withoutnoise}{$\neg$N\ }

\newcommand{\params}{\vec{\theta}}

\newcommand{\markov}{\mathcal{M}}
\newcommand{\state}{\vec{s}}
\newcommand{\tstate}{\vec{s}_t}
\newcommand{\states}{\mathcal{S}}
\newcommand{\actions}{\mathcal{A}}
\newcommand{\action}{\vec{a}}
\newcommand{\taction}{\vec{a}_t}
\newcommand{\stateinitial}{\rho_0}
\newcommand{\rewardfn}{r}
\newcommand{\transition}{\mathcal{T}}
\newcommand{\timemax}{T}
\newcommand{\policy}{\pi}

\usepackage{todonotes}

\usepackage{flushend}

\makeatletter
\let\NAT@parse\undefined
\makeatother
\usepackage[numbers]{natbib}

\title{\LARGE \bf
Leveraging Contact Forces for Learning to Grasp
}

\author{Hamza Merzi\'c$^{1}$, Miroslav Bogdanovic$^{2}$, Daniel Kappler$^{5}$, Ludovic Righetti$^{2,3}$, and Jeannette Bohg$^{2,4}$%
\thanks{The majority of this work has been conducted while all authors were with the Autonomous Motion Department at the MPI for Intelligent Systems, T\"ubingen, Germany. Part of this work was supported by New York University, the Max-Planck Society, the European Union’s Horizon 2020 research and innovation program (grant agreement No 780684 and European Research Council’s grant No 637935).}%
    \thanks{$^{1}$ DeepMind, London, UK. {\tt\small hamzamerzic@gmail.com}}
    \thanks{$^{2}$ MPI for Intelligent Systems, T\"ubingen, Germany. {\tt\small [first.lastname]@tue.mpg.de}}
\thanks{$^{3}$Tandon School of Engineering, New York University, USA. }
\thanks{$^{4}$Computer Science Department, Stanford University, USA.}
\thanks{$^{5}$Google X Robotics, Mountain View, CA, USA. {\tt\small kappler@google.com}}
}

\begin{document}

\maketitle
\thispagestyle{empty}
\pagestyle{empty}

\begin{abstract}
Grasping objects under uncertainty remains an open problem in robotics research. This uncertainty is often due to noisy or partial observations of the object pose or shape. To enable a robot to react appropriately to unforeseen effects, it is crucial that it continuously takes sensor feedback into account. While visual feedback is important for inferring a grasp pose and reaching for an object, contact feedback offers valuable information during manipulation and grasp acquisition. In this paper, we use model-free deep reinforcement learning to synthesize control policies that exploit contact sensing to generate robust grasping under uncertainty. We demonstrate our approach on a multi-fingered hand that exhibits more complex finger coordination than the commonly used two-fingered grippers. 
We conduct extensive experiments in order to assess the performance of the learned policies, with and without contact sensing. While it is possible to learn grasping policies without contact sensing, our results suggest that contact feedback allows for a significant improvement of grasping robustness under object pose uncertainty and for objects with a complex shape.
\end{abstract}

\section{Introduction}
\label{sec:intro}
Robust grasping is a fundamental prerequisite for most meaningful manipulation tasks. Yet it remains a challenge under uncertainty.
Traditional model-based approaches toward grasping use well defined metrics such as the $\epsilon$-metric~\citep{ferrari1992planning} which is based on contact points and maximum object wrenches, to determine an optimal gripper pose and configuration.
These methods typically have strong assumptions, requiring perfect knowledge of the object pose, shape and friction coefficients.
Yet, stable grasps according to $\epsilon$-metric often fail in real world executions due to imprecise control, noisy sensing or imperfect models.
Despite having a sound theoretical grounding, in real world scenarios the underlying assumptions are typically not met~\cite{balasubramanian2012}. 

Learning-based approaches~\cite{bohg2014data} attempt to alleviate these requirements to enable grasping under uncertainty about object pose, shape, type, sensor, and actuation. 
Data-driven methods toward grasping commonly have two stages. In the first stage, they infer a grasp pose and preshape from visual information; often partial and noisy 3D point clouds or RGB images. To learn the mapping from visual information to grasp pose and configuration, an annotated data set is required. These data sets are either obtained by large-scale, self-supervised data acquisition experiments, typically using two-fingered grippers~\citep{levine2016learning,pinto2016supersizing} or through simulation~\citep{kappler2015leveraging,mahler2017dex}.
In the second stage, an open loop controller executes the inferred grasp, thus, establishing contact with the object~\citep{levine2016learning,pinto2016supersizing,PasGSP17}. Approaches that adopt this two-stage, data-driven grasp planning architecture have to select the subset of grasps stable under inaccurate motion control and perceptual noise. The learned mapping has to account for all the possible uncertainty. \cite{QT-Opt} does not adopt this two-stage approach and directly learns a policy for pick and place in clutter through Q-Learning. More complex behaviors emerge, e.g. object singulation and probing. However, all these approaches employ two-fingered grippers which significantly reduces the control and learning complexity. 
Furthermore, none of these learned policies use haptic feedback despite its potential
to improve grasping behavior by providing important information about the contact interaction.
Tactile feedback is frequently used to estimate grasp stability but typically only after the grasp has already been established and before lifting the object \cite{bekiroglu2011assessing,calandra2017feeling,hausman-chebotar16iser}. 

\begin{figure}
    \centering
    \begin{minipage}{0.197\columnwidth}
        \centering
        \includegraphics[width=\linewidth]{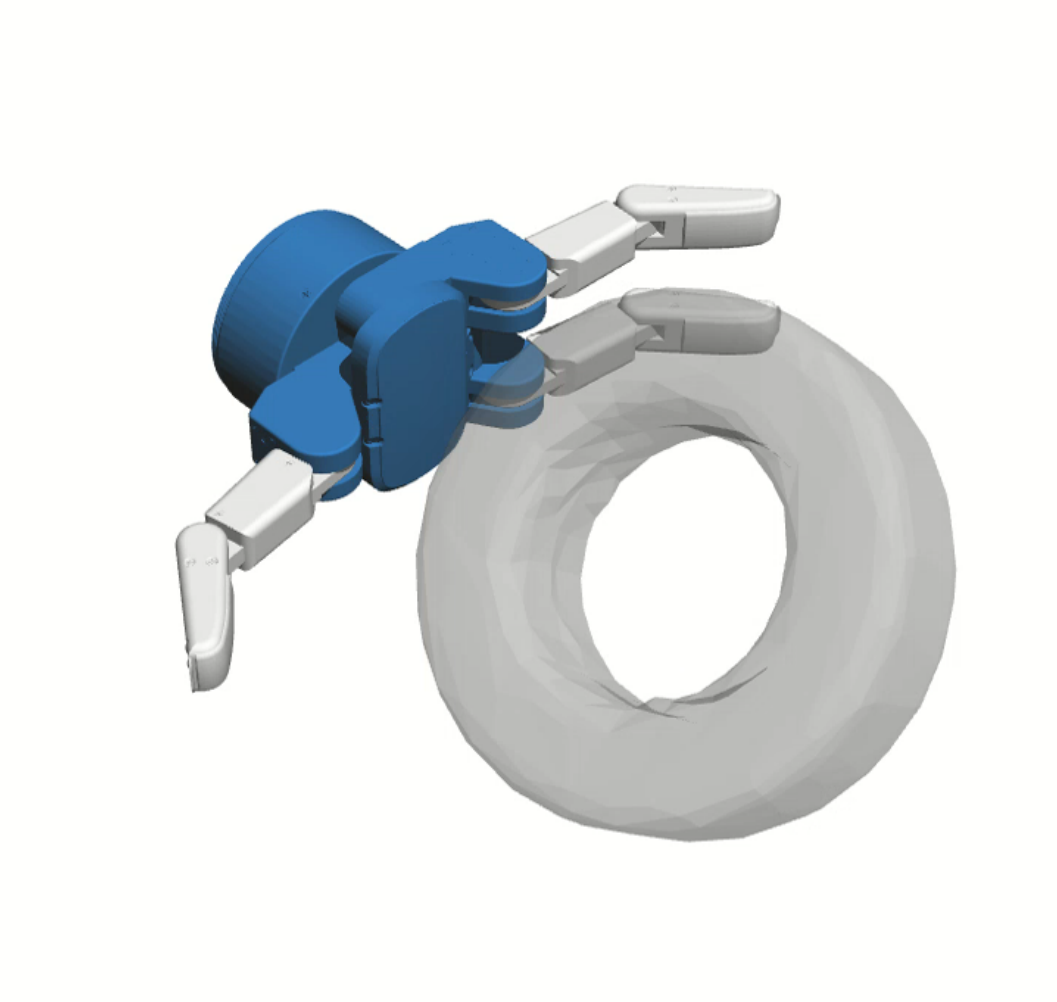} 
    \end{minipage}\hfill
    \begin{minipage}{0.197\columnwidth}
        \centering
        \includegraphics[width=\linewidth]{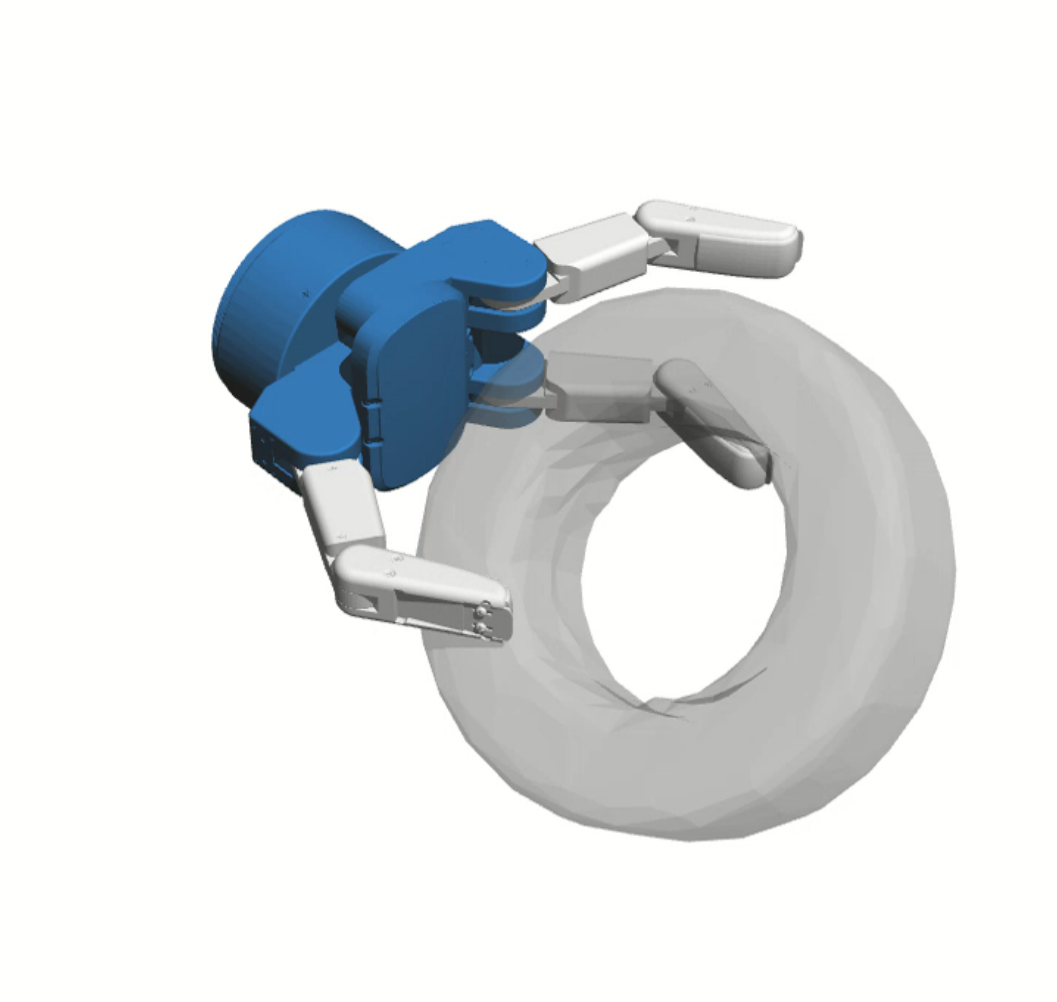} 
    \end{minipage}\hfill
    \begin{minipage}{0.197\columnwidth}
        \centering
        \includegraphics[width=\linewidth]{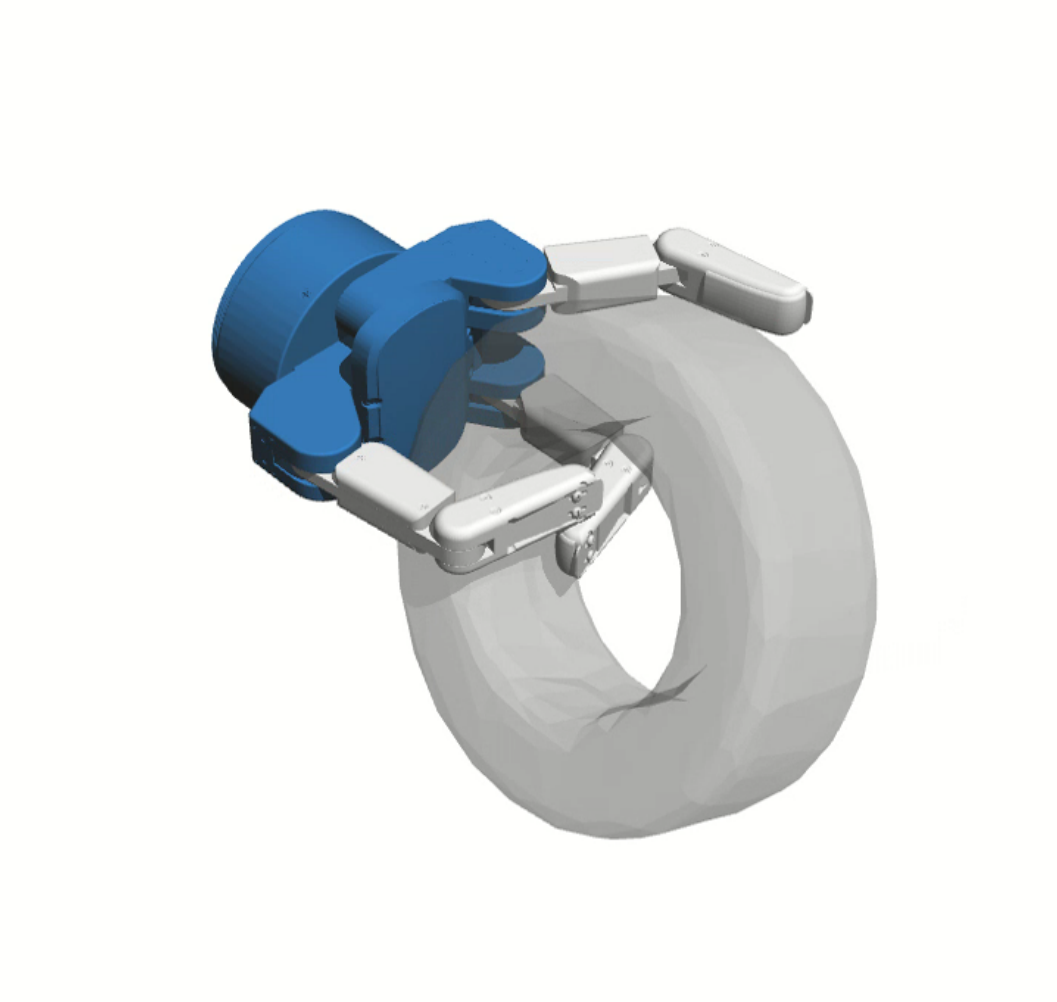} 
    \end{minipage}\hfill
    \begin{minipage}{0.197\columnwidth}
        \centering
        \includegraphics[width=\linewidth]{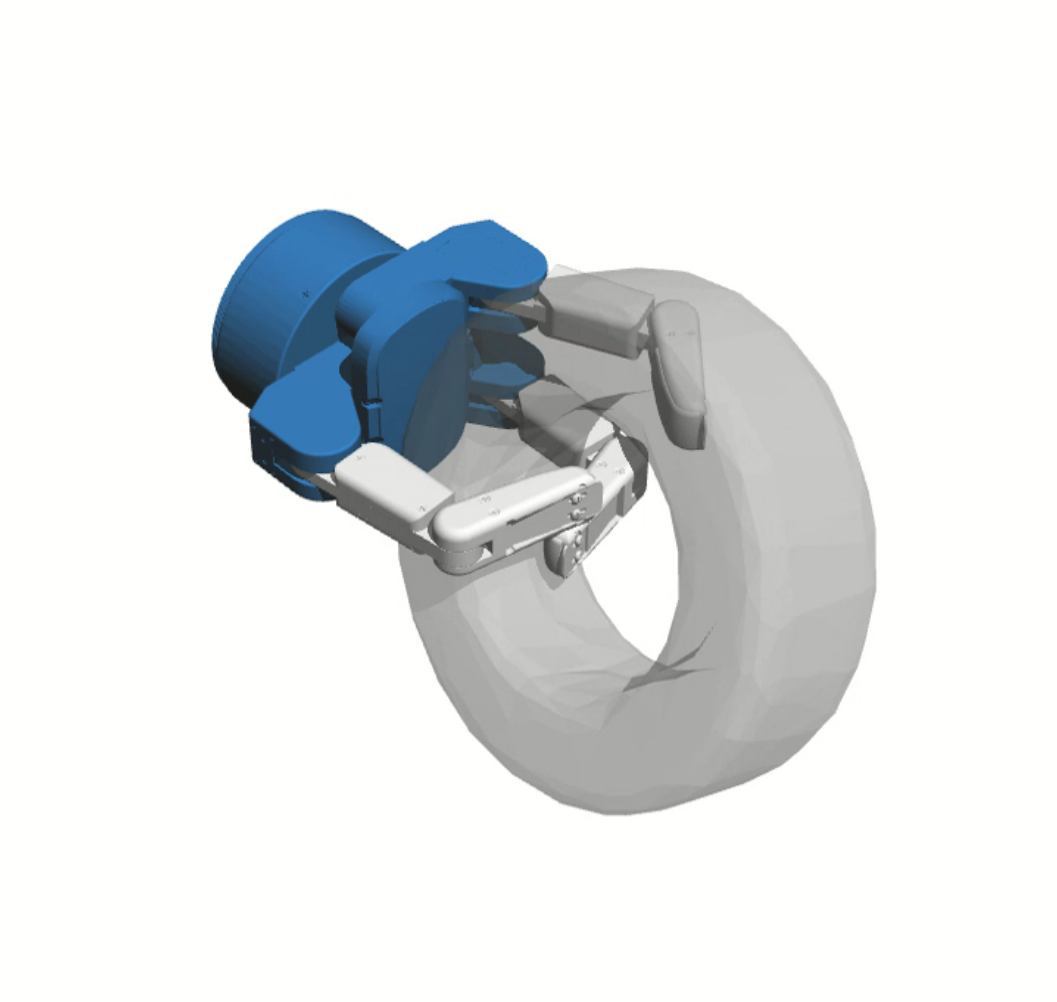} 
    \end{minipage}\hfill
    \begin{minipage}{0.197\columnwidth}
        \centering
        \includegraphics[width=\linewidth]{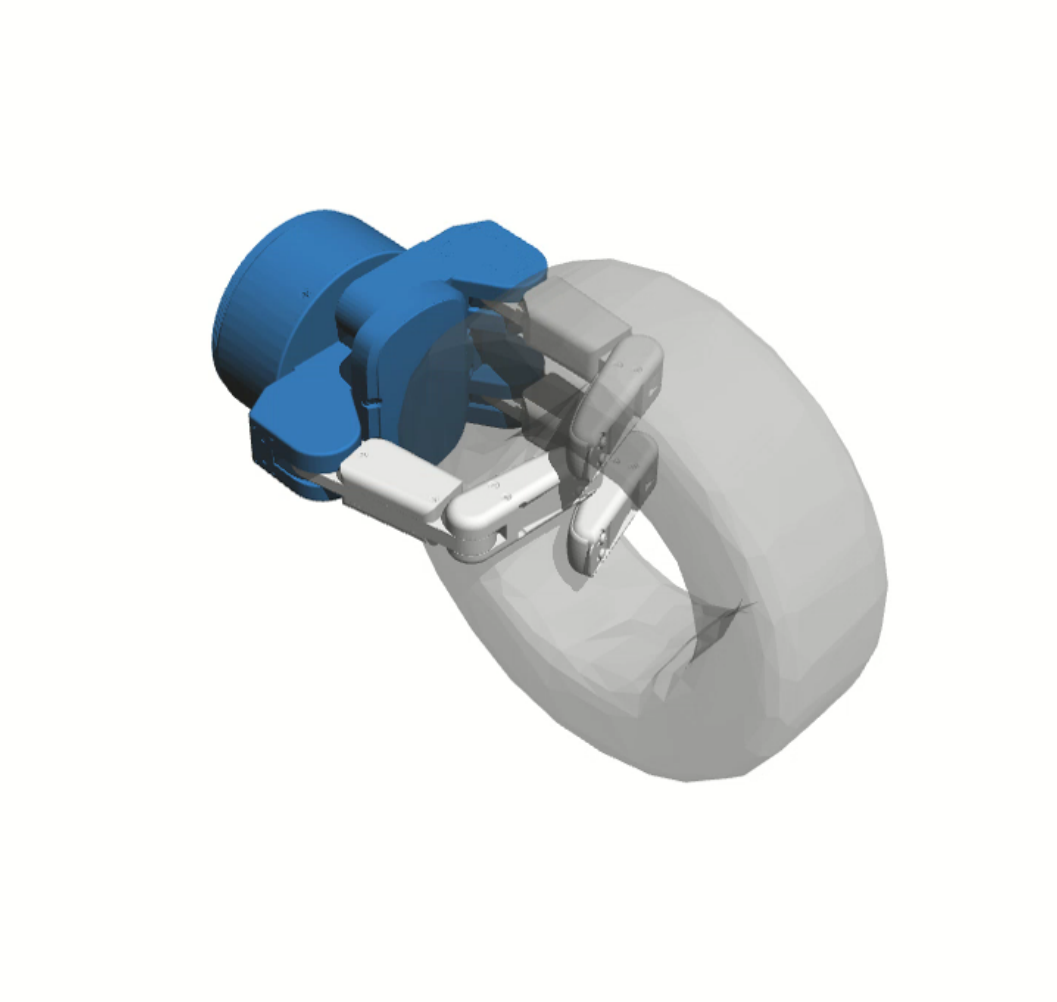} 
    \end{minipage}
    
    \begin{minipage}{0.197\columnwidth}
        \centering
        \includegraphics[width=\linewidth]{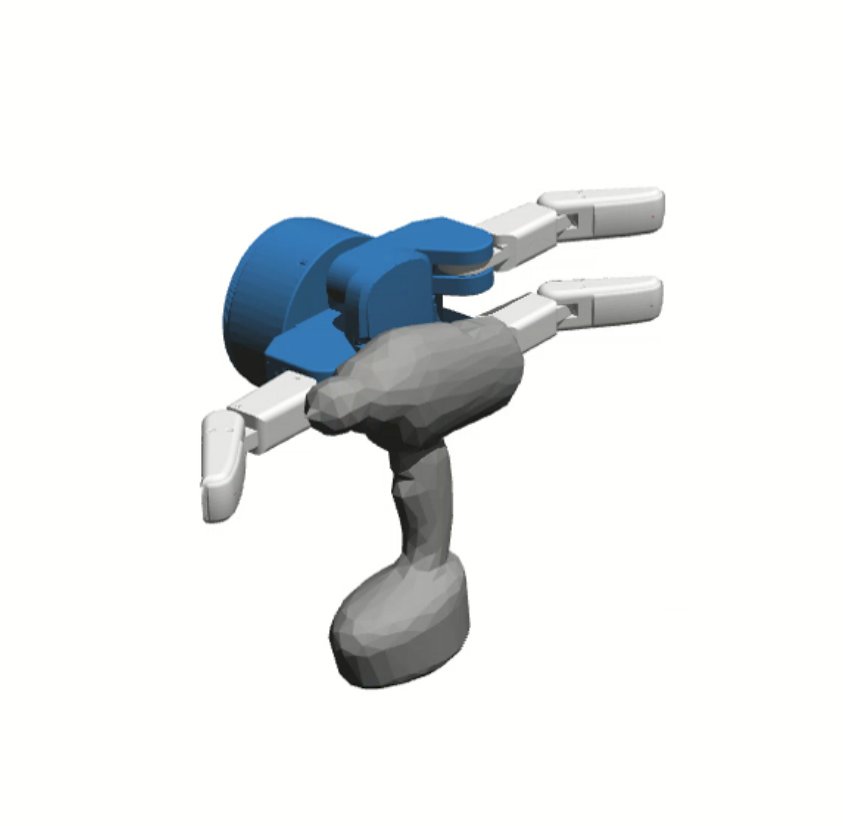} 
    \end{minipage}\hfill
    \begin{minipage}{0.197\columnwidth}
        \centering
        \includegraphics[width=\linewidth]{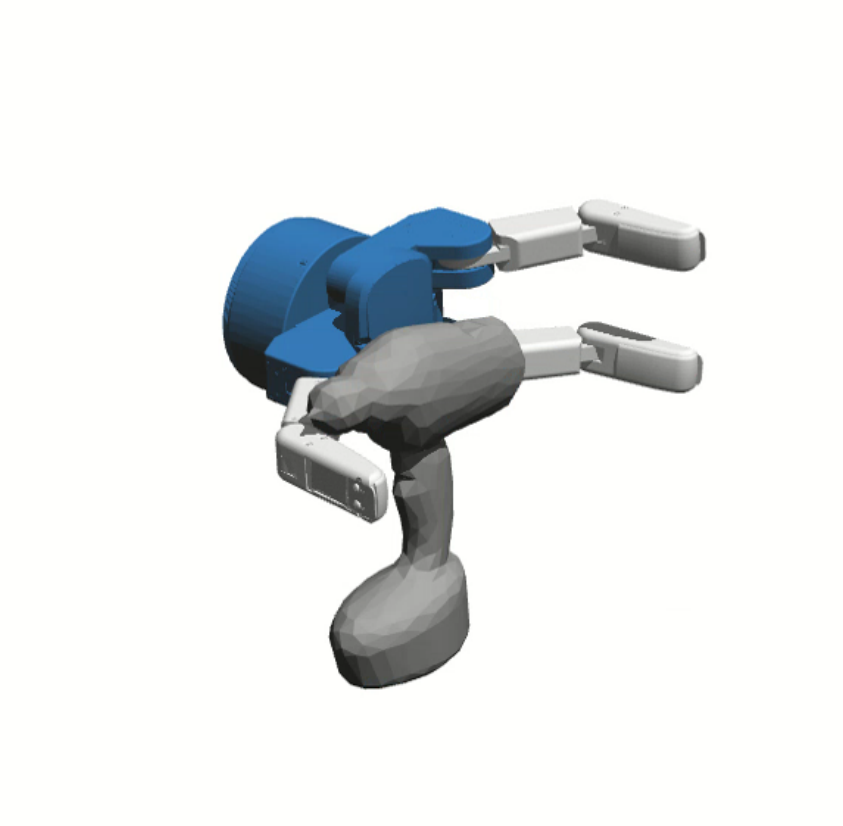} 
    \end{minipage}\hfill
    \begin{minipage}{0.197\columnwidth}
        \centering
        \includegraphics[width=\linewidth]{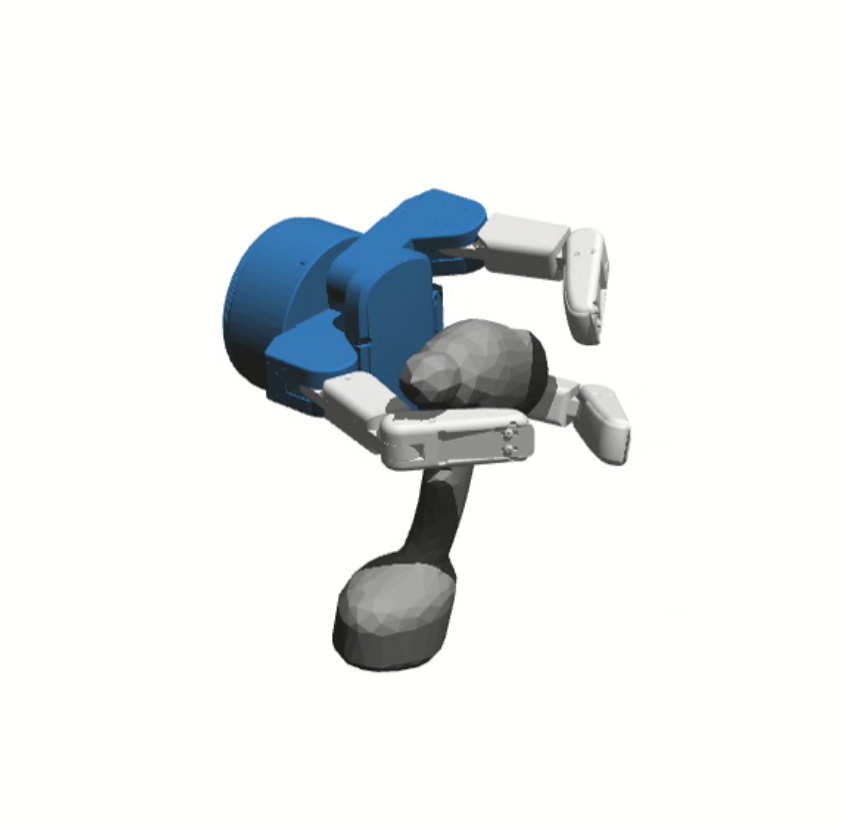} 
    \end{minipage}\hfill
    \begin{minipage}{0.197\columnwidth}
        \centering
        \includegraphics[width=\linewidth]{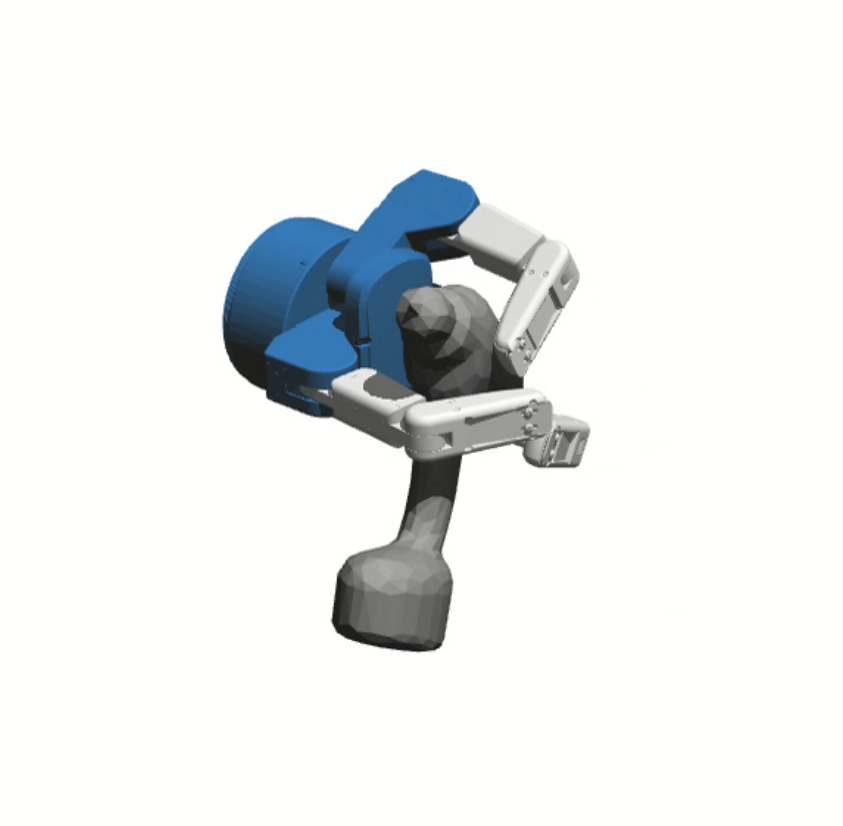} 
    \end{minipage}\hfill
    \begin{minipage}{0.197\columnwidth}
        \centering
        \includegraphics[width=\linewidth]{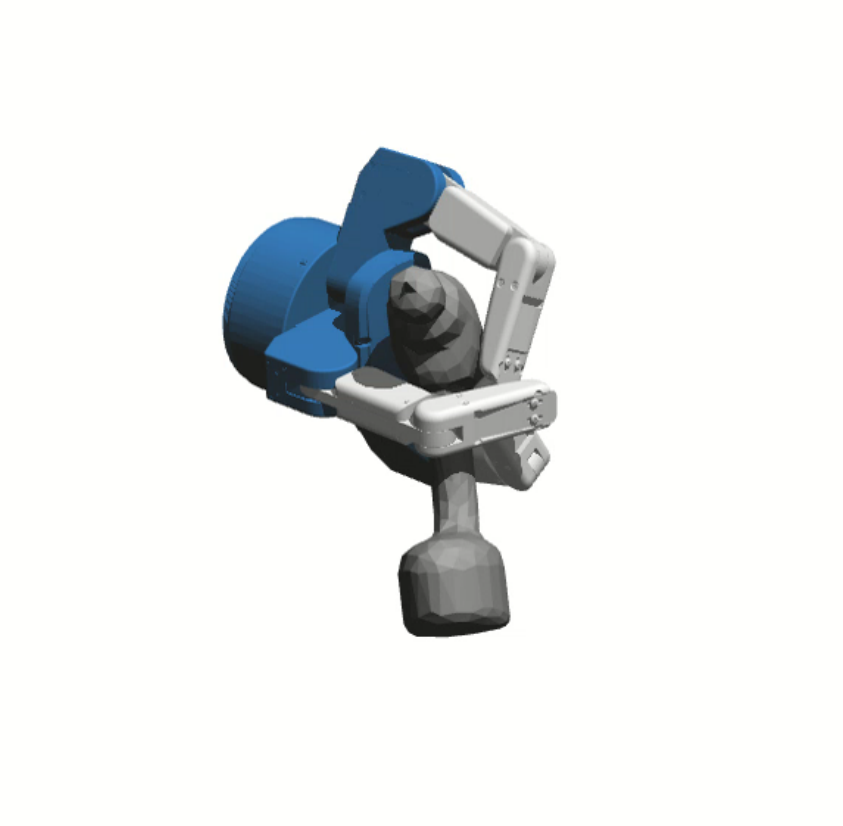} 
    \end{minipage}
    \caption{Examples of learned grasp strategies using a multi-fingered hand and contact feedback.}
\end{figure}

In this paper, we aim to demonstrate that continuous hand haptic feedback can significantly improve grasping acquisition in terms of robustness under uncertainty.
Improvement in the grasp controller will help reduce the load on the first stage of grasp planning which now only has to infer a grasp pose for a robust {\em closed-loop\/} controller instead of an {\em open-loop\/} controller more sensitive to uncertainty. 

Recent advances in deep reinforcement learning (DRL) have shown promising results on problems that were previously rendered intractable due to high-dimensional state-action spaces~\citep{mnih2013playing,schulman2017proximal}. Motivated by these results, we learn contact-feedback policies with model-free DRL for grasping known objects.
This alleviates contact and sensor modeling required for optimal control formulations. Different from the majority of recent learning-based approaches for grasping, we use a multi-fingered gripper, enabling richer grasping strategies.  Different from recent learning-based approaches for dexterous manipulation with multi-fingered hands \citep{kumar2016learning,rajeswaran2017learning}, we assume feedback from contact sensors and noisy object pose estimates e.g. from visual object trackers~\cite{wuthrich-iros-2013}. Improving robustness in this scenario is crucial since uncertainty on object pose is inevitable in dynamically changing, potentially cluttered environments due to partial observation and sensor noise.

The main contributions of our work are twofold:
(i) We present empirical evidence that it is possible to learn grasp acquisition policies with DLR using contact feedback,
(ii) our experimental evaluation suggests that contact sensing is beneficial for learning grasp policies. These policies exhibit improved robustness to noisy object pose estimates.

In the following we present related work in Section~\ref{sec:related_work}.
We then state the general problem formulation and learning approach in Section~\ref{sec:problem},
followed by a detailed introduction of our learning environment in Section~\ref{sec:environment}.
We then present empirical results in simulation in Section~\ref{sec:evaluations} and conclude in Section~\ref{sec:conclusions}.

\section{Related Work}
\label{sec:related_work}
%
Robust and stable grasping is essentially a matter of controlling contact forces.
Therefore, more realistic contact modeling and contact interaction planning is an interesting research direction toward grasping. 
While such approaches have been successfully applied for locomotion planning \cite{Ponton:2018vy,Posa:2013ez,tonneau18} they still
result in challenging optimization problems. They are also
very sensitive to model uncertainty (e.g. object pose and contact locations).
Data-driven approaches towards grasping have been prominent in the field of robotic manipulation research \citep{bohg2014data}.
Hereafter, we focus on reviewing model-free data-driven approaches for tasks which require complex contact interactions to learn feedback policies for grasping.

Recent advancements in deep learning, large-scale real world grasp experiments and guided policy search have shown impressive results for manipulation tasks. \citet{levine2016end} presented first results on end-to-end closed-loop policy learning for contact-rich manipulation tasks without considering contact information but only RGB images and joint encoder readings as input.
This work differs from our approach in two aspects. 
First, no contact feedback is used for learning the feedback policy.
Second, the underlying learning approach is fundamentally different, effectively iterating between local dynamic model learning, optimal control, and policy learning. We employ a reward-based model-free deep reinforcement learning method.
Two large-scale self-supervised grasping experiments~\cite{levine2016learning,pinto2016supersizing} have shown that it is possible to learn a successful grasping pipeline from scratch, consisting of a closed-loop vision-based policy and an open-loop grasp acquisition controller. 
Both experiments use simple 2-fingered gripper, greatly reducing the grasp controller complexity, and learn a grasp success predictor~\cite{levine2016learning} or a grasp pose predictor~\cite{pinto2016supersizing} whereas we learn a closed-loop grasp acquisition controller. 

Recent results in DRL~\cite{mnih2013playing} have shown how RL agents can be trained from raw, nonlinear and high-dimensional observations.
DLR has been successfully applied to dexterous manipulation tasks such as pivoting and grasping with multi-fingered hands  \cite{kumar2016learning,rajeswaran2017learning}.
Yet, to the best of our knowledge despite requiring complex contact interactions no DLR approach so far has been using contact sensor information for policy learning.
\citet{antonova2017reinforcement} and \citet{peng2017} demonstrated that injecting noise in different parts of the simulation during policy learning, alleviates problems typically observed when transferring policies from simulation to the real world. 
Different to this work, we present empirical results in simulation that show that including contact information in the state improves the performance under uncertainty. 

Due to the importance of contact feedback for stable grasping, a lot of work has been done to integrate tactile sensors in grasping frameworks.
For example, tactile sensors have been used for estimating grasp stability \cite{bekiroglu2011assessing, su2015force} and for adapting the exerted forces at contact points, e.g. by detecting slip~\cite{veiga2015stabilizing}.
\citet{hausman-chebotar16iser} presented successful re-grasping policies based on a multi-stage grasping pipeline, consisting of grasp success prediction, re-grasping policy learning and supervised policy fusion.
Different to our approach, contact sensing is used for grasp-success prediction and learning new re-grasp candidates.
Similarly, \citep{calandra2017feeling} has recently proposed an end-to-end approach for predicting grasp outcomes by combining visual and tactile information, based on an optical tactile sensor~\citep{dong2017improved}. 
Also this approach significantly differs from our methodology since the learned grasp outcome predictor is used in a model predictive control framework to achieve a stable grasp, potentially requiring re-grasping. We directly learn a feedback policy using contact information for grasp acquisition.

\section{Problem Statement and Approach}
\label{sec:problem}
We propose to learn a control policy for grasp acquisition with a three-fingered hand. We aim to investigate the role of contacts for policy learning and robustness improvements with respect to object pose uncertainties. We use model-free reinforcement learning to learn a policy for grasp acquisition, thus avoiding the need for accurate contact and dynamic models which are hard to define or learn in contact-rich scenarios with hybrid dynamics and in the presence of uncertainty.

We phrase the problem of grasp acquisition as a finite-horizon discounted Markov decision process (MDP) $\markov$, with a state space $\states$, an action space $\actions$, state transition dynamics $\transition : \states \times \actions \to \states$, an initial state distribution $\stateinitial$, a reward function $\rewardfn : \states \times \actions \to \mathbb{R}$, horizon $\timemax$, and discount factor $\gamma \in (0, 1]$.
Hence, we are interested in maximizing the expected discounted reward 
\begin{equation}
J(\policy) = \mathbb{E}_\policy \left[\sum^{\timemax-1}_{t=0} \gamma^t \rewardfn(\tstate, \taction) \right]
\label{eqn:loss}
\end{equation}
to determine the optimal stochastic policy $\policy : \states \to \mathbb{P}(\actions)$. We parameterize this policy with learnable parameters $\params$:
\begin{equation}
\policy(\action|\state) = \policy_{\params}(\action|\state)
\end{equation}
One way to optimize $J(\policy_{\params})$ over $\params$ is by directly using gradient ascent. This approach is known as the Policy Gradient method \cite{sutton1999} and it demonstrates that the value of the gradient can be expressed as:
\begin{equation}
\nabla_{\params} J(\policy_{\params}) = \mathbb{E}_{\policy_{\params}} \left[\nabla_{\params} \log \policy_{\params}(\action|\state) {Q(\state, \action)} \right],
\end{equation}
where $Q(\state, \action)$ is the state-action value function.
This formulation illustrates how to use the samples obtained from policy rollouts to estimate this gradient and perform gradient ascent.

In practice there are several algorithms that give significantly better results than the basic policy gradient implementation. 
One of them, which is used for all evaluations presented in this paper, is Trust Region Policy Optimization (TRPO) \cite{schulman2015trust}.
TRPO optimizes the following objective function, subject to a constraint on the change in the policy update:
\begin{align*}
    \max_{\widetilde{\params}} &\qquad  \mathbb{E}\left[\frac{\policy_{\widetilde{\params}}(\action|\state)}{\policy_{\params}(\action|\state)}A_{\params}(\state, \action)\right] \\
    \text{s.t.} & \qquad \mathbb{E}\left[D_{KL}(\policy_{\params}(\cdot|\state)||\policy_{\widetilde{\params}}(\cdot|\state))\right] \le \delta
\end{align*}

\noindent where $A(\state, \action)$ is the advantage function defined as $A(\state, \action) = Q(\state, \action) - V(\state, \action)$.

The policy is represented as a multivariate Gaussian policy with diagonal covariance, which also constitutes the learning parameters. In both the advantage function estimate as well as the policy, we use a neural network with three hidden layers of 64 units each followed by a tanh nonlinearity. 
The generalized advantage estimation is not used, and we use the default values for all remaining open hyper-parameters according to the OpenAI baseline~\cite{baselines} implementation.

\section{Policy Learning for Grasp Acquisition}
\label{sec:environment}
In this section, we propose a way to include contact-feedback for learning a grasp acquisition policy for known objects. We first define the state and action space. Then we introduce our simulated learning environment, grasping episodes and finally reward design.

\subsection{Object Models and Robot Hand}
\begin{figure}
    \centering
    \includegraphics[width=\linewidth]{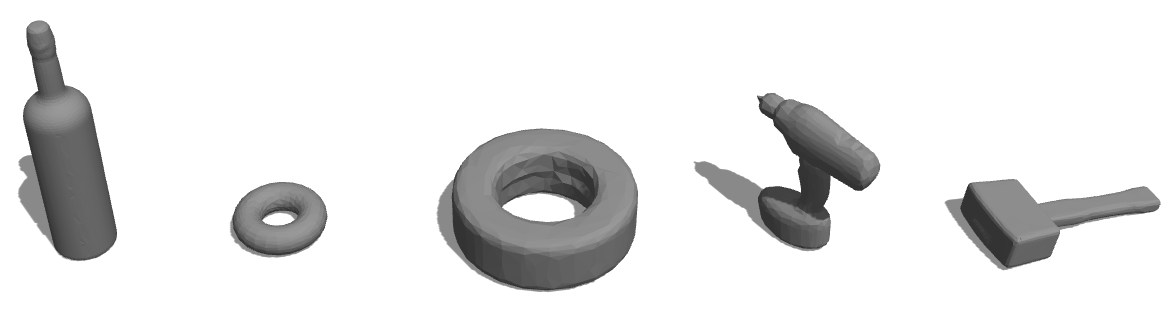}
    \caption{Five objects from the grasping database~\citep{kappler2015leveraging}. From left to right: bottle, donut, tape, drill, hammer.}
    \label{fig:objects}
\end{figure}

\citet{kappler2015leveraging} presented a database of object mesh models that are each associated with on average 500 pre-grasps from which a grasp acquisition controller can be initialized. In this paper, we use five of these objects varying in size and shape complexity (see Fig.~\ref{fig:objects}): $\objects= \{\text{donut}, \text{bottle}, \text{hammer}, \text{drill}, \text{tape}\}$. The set of pre-grasps per object is referred to as $\mathcal{P}_o$

\begin{figure}
    \centering
    \includegraphics[width=0.8\linewidth]{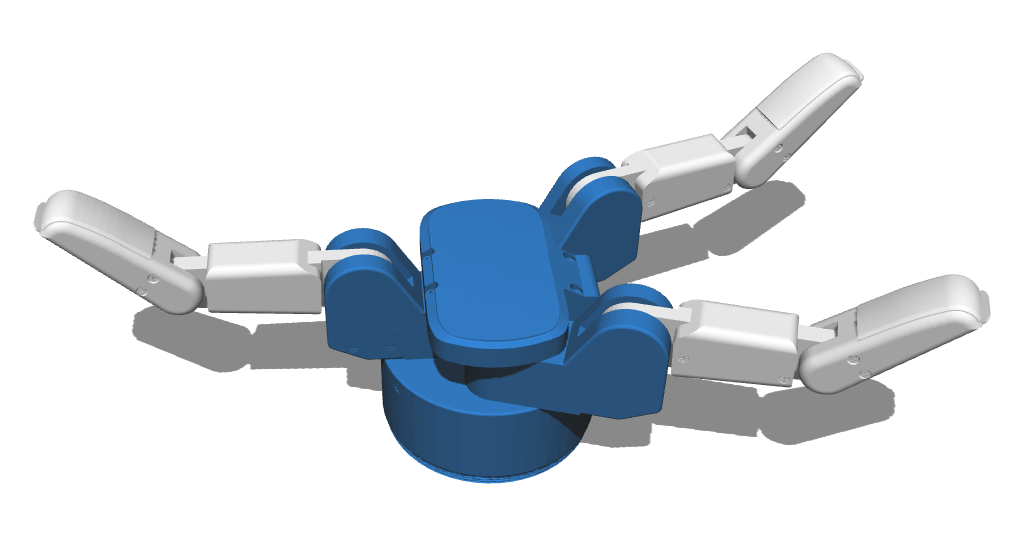}
    \caption{The BarretHand with four independent DoF and three links per finger. Each DoF can be torque-controlled. For each link we receive contact feedback in form of a 3D force vector.}
    \label{fig:barrett}
\end{figure}

The hand used in \citep{kappler2015leveraging} and in our experiments is the BarretHand (see Fig.~\ref{fig:barrett}), a three-fingered hand frequently used both in industry and in academia. It has four independent degrees of freedom (DOF),  three of which are in charge of opening and closing the fingers and the fourth one controlling the spread between the left and right fingers.

\subsection{State and Action Space}
In the following we will compare policies learned with and without contact feedback in order
to evaluate its benefits. Each of these variants uses a nominal state $\states_n = \{\enc, \pose, \twist \}$ where $\enc$ refers to the 4 dimensional joint measurements of the robot hand, $\pose$ refers to the 7 dimensional object pose (3 for position and 4 for orientation in the form of a quaternion) and $\twist$ refers to the 6 dimensional object twist (3 for each, linear and angular velocity). For policies that get contact feedback, $\states$ also include simulated contact force measurements $\cts$, i.e. $\states = \states_n \cup \{\cts\}$. In our simulations, contact sensors provide contact force vectors acting on each link in the inertial frame of the hand. This yields a 27-dimensional vector of contact forces additional to the 17 dimensional nominal state $\states_n$.
The robot hand is torque-controlled to allow for finger impedance regulation and compliant manipulation. Therefore, an action $\action$ is a four-dimensional vector containing joint torques.

\subsection{Simulated Learning Environment}
We developed a learning environment built on top of Gazebo \cite{koenig2004design} using the ODE physics engine back end, which simulates hard contacts. This is in contrast to other popular simulators such as MuJoCo~\citep{Todorov2012MuJoCoAP} that have soft contact models which may be less realistic for simulating manipulation tasks with complex contact interactions.
The learning environment is equipped with a database of objects, pre-grasp configurations and grasping metrics as defined in~\citep{kappler2015leveraging}.

\subsection{Grasping Episode}
\begin{figure}[t]
    \centering
    \includegraphics[width=\linewidth]{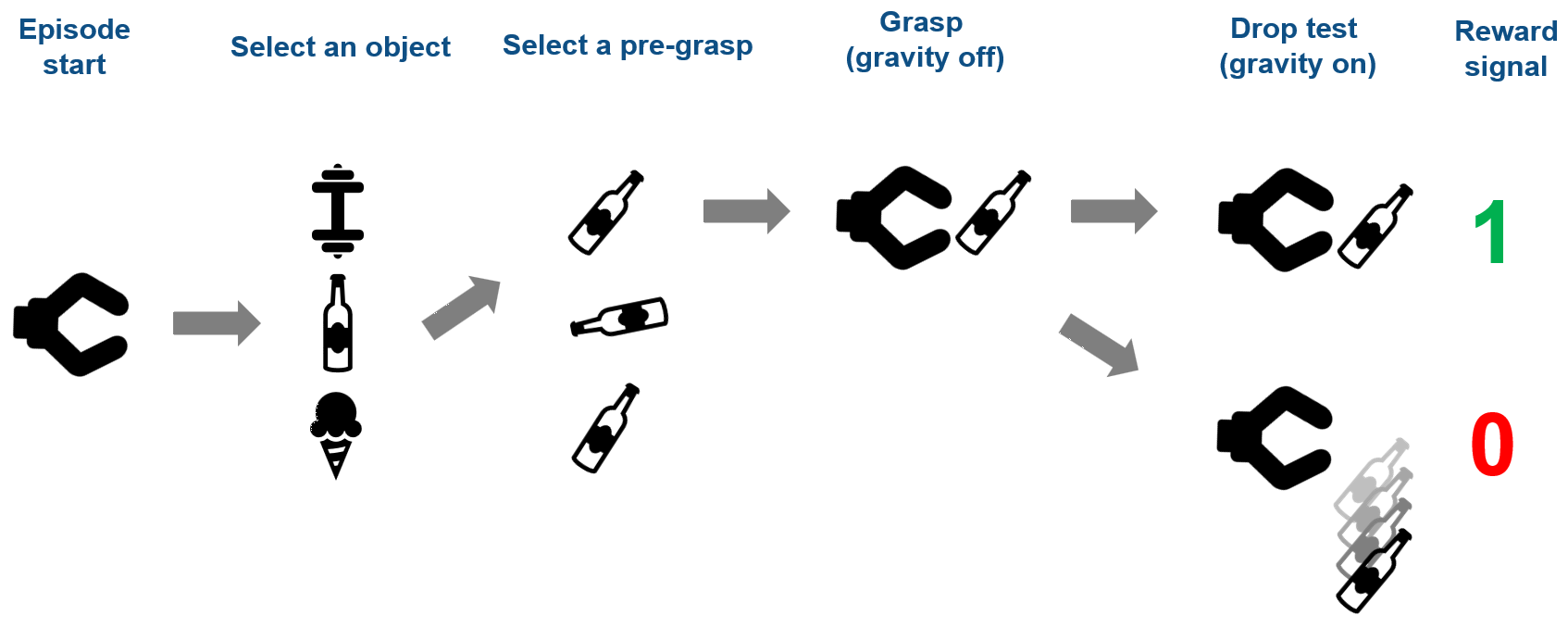}
    \caption{Breakdown of a single grasping episode.}
    \label{fig:episode}
\end{figure}

A single grasping episode for which we are trying to find an optimal policy is visualized in Fig.~\ref{fig:episode}. Note that during an episode, the base of the hand remains fixed as we are only controlling the fingers. Also gravity is disabled throughout the episode. One episode is constructed as follows:

\begin{enumerate}
\item Randomly select a pair of object and pre-grasp: $(\object, \pregrasp)$, $\object \in \objects$, $\pregrasp \in \pregrasps$.
\item Place the hand $\gripper$ and the object $\object$ in the positions defined by the pre-grasp configuration $\pregrasp$. 
\item Roll-out the current policy $\policy$ in this environment.
\item Perform a {\em drop test\/} after $\timemax$ time steps and assign the corresponding reward.
\end{enumerate}
During the \textit{drop test}, a 12N force is applied on the object; first in the upward and then in the downward direction, both for the duration of 0.5s. If the object remains within a 5cm diameter of its position prior to the \textit{drop test}, we consider the test successful, otherwise a failure. The corresponding binary reward is returned. In all of the experiments the time horizon $T$ equals 1000. A single simulation step takes $10$ ms, making the whole episode last $10$s of simulation time.

\subsection{Reward design}
A common challenge, especially in robotics, is designing a reward function which reflects the task at hand, but is also carefully shaped to guide the exploration and increase learning speed and stability \cite{ng1999policy}. We use simple grid search and a classical reward shaping approach to obtain the multimodal reward. The final reward is a linear combination of six signals with corresponding weights $\alpha_i$, $i \in \{1, 2, \dots 6\}$:
\begin{align*}
r & = \alpha_1 \, \Delta n_{cont} + \alpha_2 \, \Delta||\pregrasp - \pose||_2 + \alpha_3 \, ||\action||^2_2  \\ 
& +  \alpha_4 \, ||\twist||^2_2  + \alpha_5 \, \Delta r_{dist\_fingertips}  + \alpha_6 \, r_{drop\_test}
\end{align*}
Here, $\Delta n_{cont}$ is the change in the total number of links in contact with the object,  $\pose$ and $\pregrasp$ are the positions of the object's center of mass and the base of the gripper, respectively, thus the term represents the change in distance of the object to the hand.
The third and fourth term are regularization terms applied to the joint torques $\action$ and the object's linear velocity $\twist$.
The term $\Delta r_{dist\_fingertips}$ quantifies the change in the mean distance of the fingertips of the hand to the object, which we use to guide the learning towards closing the hand and getting in contact with the object.
The final term is the previously discussed drop test reward which constitutes the largest part of the total reward.

\section{Evaluations}
\label{sec:evaluations}
In the following we present our experimental evaluation, designed to answer the following questions:
\begin{enumerate}
\item Can we learn policies which successfully perform grasp acquisition?
\item Does contact feedback improve grasp policy learning?
\item Can we generalize to unseen pre-grasp configurations by using contact feedback?
\item Can we learn policies robust to pose estimation noise when considering contacts?
\end{enumerate}

\begin{figure}[thpb]
    \centering
    \includegraphics[width=\linewidth]{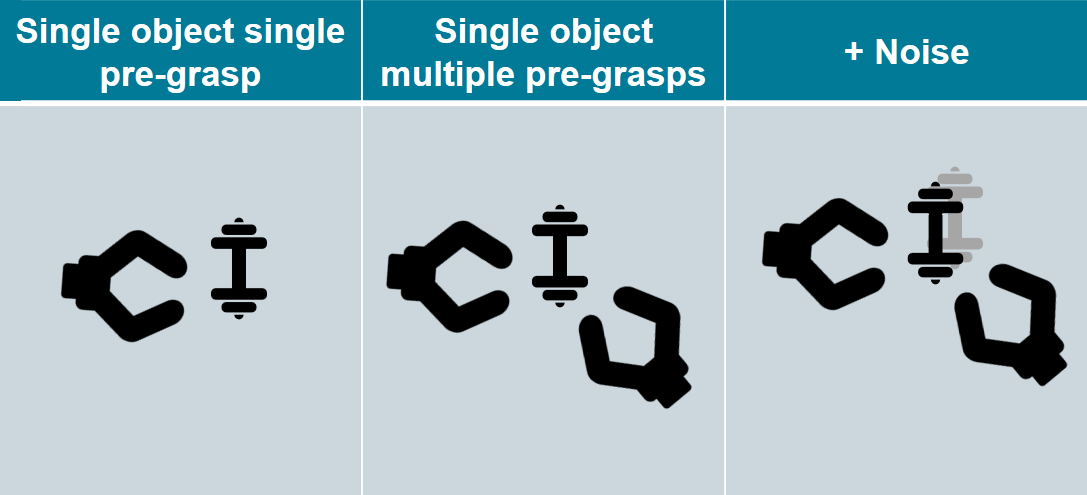}
    \caption{Visualization of the experimental categories.}
    \label{fig:experiments}
\end{figure}

The remainder of the evaluation is organized as follows.
In Section~\ref{subsec:metrics} we present the evaluation metric used across all of the experiments. 
The two baselines, used for comparison are described in Section~\ref{subsec:baselines}.
Thereafter, we discuss three experimental settings (Fig.~\ref{fig:experiments}), addressing the previously posed questions.
We use the same object set $\objects= \{\text{donut}, \text{bottle}, \text{hammer}, \text{drill}, \text{tape}\}$, shown in Figure~\ref{fig:objects}, for all experiments.

\subsection{Evaluation Metric}
\label{subsec:metrics}
A single grasp problem instance consists of a pre-grasp $\pregrasp \in \pregrasps$ and the corresponding
object $\object \in \objects$. 
We split the resulting data set into a disjoint train (70\%) and test (30\%) set, if not stated otherwise.
The grasp policy is optimized on the training set and the resulting policy is evaluated once on every sample of the test set.
An evaluation on a test example is considered stable if the grasped object withstands the earlier described droptest.
We report all the results as percentages of stable grasp attempts on the test set.

\subsection{Baselines}
\label{subsec:baselines}
Additional to comparing policies with and without contact feedback, we also introduce two non-learning baselines: 
\begin{enumerate}
    \item An open-loop policy applying constant torques to each controllable joint during the whole episode.
    \item Physics-based grasp stability metric~\citep{kappler2015leveraging} per pre-grasp.
\end{enumerate}
The physics-based metric for a pre-grasp is also generated by an open-loop grasp controller but differs in two important aspects from the constant torque baseline. First, the final metric per pre-grasp is the successful percentage of 31 grasp trials on noisy object pose estimates. Here, successful means that the object has not slipped out of the hand during grasping. Second, the simulation environment only resolves contacts after all fingers have made contact or are fully closed. We consider a pre-grasp successful according to the physics score if it is larger than 0.5, i.e. the majority of the grasp trials was successful.

\subsection{Single object single pre-grasp}
\label{subsec:single}

The goal of this evaluation is to test the benefits of including contact sensor modalities for learning how to grasp.
To that end we apply the learning algorithm individually on 100 randomly selected pre-grasps for each of the test objects, obtaining a policy per pre-grasp.
We assume perfect knowledge of the object's pose as commonly done in related work on learning multi-fingered manipulation, e.g.~\citep{kumar2016learning}.
The results, along with the baselines, are shown in Table~\ref{tab:single}.

\begin{figure}[thpb]
    \centering
        \begin{tabular}{l|rrrr}
            Objects              &
            \multicolumn{1}{c}{\begin{tabular}[c]{@{}c@{}}\physicsscore \textgreater \, 0.5 \end{tabular}} &
            \multicolumn{1}{c}{\begin{tabular}[c]{@{}c@{}}\constanttorque \end{tabular}} &
            \multicolumn{1}{c}{\begin{tabular}[c]{@{}c@{}}\withoutcont \end{tabular}} &
            \multicolumn{1}{c}{\begin{tabular}[c]{@{}c@{}}\withcont \end{tabular}} \\ \hline

            donut  & 37\% & 57\% & 81\% & \textbf{91\%} \\

            bottle & 31\% & 43\% & 58\% & \textbf{62\%} \\

            hammer & 26\% & 27\% & 45\% & \textbf{48\%} \\

            drill  & 15\% & 22\% & 38\% & \textbf{46\%} \\

            tape   & 17\% & 28\% & 35\% & \textbf{37\%}
        \end{tabular}
    \captionof{table}{Single object single pre-grasp. \physicsscore stands for the physics score, \constanttorque for the constant-torque policy. Symbols \withcont and \withoutcont denote policies that were trained with and without contacts, respectively.}
    \label{tab:single}
\end{figure}

Each entry corresponds to a percentage of successful pre-grasps.
As we can see in Table~\ref{tab:single}, there is a substantial improvement when incorporating feedback compared to the open loop policies.
Including contact feedback further improves the grasping performance on each of the tested objects.
This is the only experiment, where the policies are tested on the same pre-grasps they were trained on. In the other experiments, the train and test set are distinct as we do not only test the performance of a policy on a given set, but its generalization capabilities as well. 
Therefore, results from this subsection can also be used as a reference value on the number of graspable pre-grasps. 

\subsection{Single object multiple pre-grasps}
\label{subsec:multiple}

Evaluations presented in this subsection are designed to test generalization to previously unseen pre-grasps on the same object.
We do this by splitting the pre-grasps into a train and test set as described earlier, and report the results on the test set in Table~\ref{tab:multiple}.
Unlike in the previous subsection where we obtain a policy for each pre-grasp, here we obtain a policy for each object.

\begin{figure}[thpb]
    \centering
        \begin{tabular}{l|rrrr}
            Objects              &
            \multicolumn{1}{c}{\begin{tabular}[c]{@{}c@{}}\physicsscore \textgreater \, 0.5 \end{tabular}} &
            \multicolumn{1}{c}{\begin{tabular}[c]{@{}c@{}}\constanttorque \end{tabular}} &
            \multicolumn{1}{c}{\begin{tabular}[c]{@{}c@{}}\withoutcont \end{tabular}} &
            \multicolumn{1}{c}{\begin{tabular}[c]{@{}c@{}}\withcont \end{tabular}} \\ \hline

            donut  & 48.0\% & 48.3\% & 51.7\% & \textbf{56.3\%} \\

            bottle & 22.1\% & 51.5\% & 67.7\% & \textbf{69.1\%} \\

            hammer & 26.4\% & 34.9\% & 35.8\% & \textbf{38.7\%} \\

            drill  &  7.3\% & 20.7\% & 23.2\% & \textbf{25.6\%} \\

            tape   & 15.6\% & 33.3\% & 40.0\% & \textbf{41.5\%}
        \end{tabular}
        \captionof{table}{Results for the single object multiple pre-grasps experiments. We use the same notation as in Table \ref{tab:single}} 
    \label{tab:multiple}
\end{figure}

Again, we can conclude that feedback provides a substantial performance improvement compared to the open-loop policies. 
Additionally, for all objects we have observed a slight performance improvement when including contact feedback as input to the policy. 
Compared to Table~\ref{tab:single}, we can see a general decrease of grasp performance as the policies now have to generalize to unseen pre-grasps. On two objects grasp performance is improved. However, the pre-grasp in the test set also seem to be easier as indicated by the higher percentage of the constant-torque policy.

\subsection{Multiple pre-grasps with noise in pose estimation}
\label{subsec:with_noise}

The goal of this subsection is to test how adding noise affects grasping performance in terms of generalization.
The evaluation setup is similar to the one in the previous subsection, with the only difference being the added noise to the object's pose estimation.
At every time step the policy receives a noisy object pose estimate where the noise is added in both position and orientation.
The position noise is a uniform random variable within a sphere of radius $1.5\text{cm}$, while the orientation noise is represented as angle-axis rotation with the axis selected uniformly at random and the angle is uniform random with magnitude of $0.3\text{rad}$.
The results are shown in Table~\ref{tab:with_noise}.
The first two columns are the same as in the previous evaluation.
The second two columns show the results of policies trained without noise ($\neg$N), while in the last two columns results of policies trained with noise (N) are shown, both with (C) and without ($\neg$C) contacts.
All of the evaluations were performed with noise on a held-out test set of pre-grasps.

\begin{figure}[thpb]
    \centering
    \resizebox{\columnwidth}{!}{%
        \begin{tabular}{l|rrrrrr}
            Objects              &
            \multicolumn{1}{c}{\physicsscore \textgreater \, 0.5} &
            \multicolumn{1}{c}{\constanttorque} &
            \multicolumn{1}{c}{\withoutcont\withoutnoise} &
            \multicolumn{1}{c}{\withcont\withoutnoise} &
            \multicolumn{1}{c}{\withoutcont\withnoise} &
            \multicolumn{1}{c}{\withcont\withnoise} \\ \hline

            donut  & 37.9\% & 48.3\% & 50.6\% & 51.7\% & 52.9\% & \textbf{56.3\%} \\

            bottle & 27.9\% & 51.5\% & 64.7\% & 64.7\% & 63.2\% & \textbf{70.6\%} \\

            hammer & 16.0\% & 34.9\% & 33.9\% & 35.8\% & 34.9\% & \textbf{39.6\%} \\

            drill  & 7.3\%  & 20.7\% & 18.3\% & 19.5\% & 19.5\% & \textbf{28.0\%} \\

            tape   & 15.5\% & 33.3\% & 39.3\% & 40.7\% & 43.7\% & \textbf{44.4\%}
        \end{tabular}
    }
    \captionof{table}{Results of the multiple pre-grasps with noise experiments. We use the same notation as in Table~\ref{tab:single}. In addition, \withnoise and \withoutnoise denote policies that were trained with and without noise, respectively.}
    \label{tab:with_noise}
\end{figure}

If we compare the policies that are trained with pose noise but are either using or not using contact feedback (\withcont\withnoise versus \withoutcont\withnoise), we observe a significant performance increase when including contact feedback.
The highest changes in performance can be observed for the drill.
Our hypothesis is that this is due to the complex shape of the drill when compared to the other objects. Thus having contact feedback in noisy scenarios plays a crucial role.

In general, policies that are trained on noisy pose estimates show an improved performance on the test set.
Comparing the results to results of the previous subsection, we notice a small performance improvement, even though here we include noise in the evaluations.
This demonstrates that including noise during training improves generalization capabilities of the learning agents.

\section{Conclusions}
\label{sec:conclusions}
In this paper, we used model-free reinforcement learning for multi-fingered grasping of known objects. Different from previous work on learning policies for dexterous manipulation, we include feedback from contact sensors and assume noisy object pose estimates. As expected, we observed that feedback policies outperform commonly employed open-loop controllers. We also showed how feedback from contact sensors improves the robustness of grasping specifically under object pose uncertainty through noisy sensing. This is particularly emphasized when the object shape is complex. And finally, we showed how training under noisy conditions improves the robustness of a grasping policy even when not using feedback from contact sensing.

As future work, our goal is to execute these policies on a real robot equipped with the same hands. This would be an extension of our dynamic manipulation system~\citep{kappler2018} that would allow it to perform a larger set of grasps on the visually tracked objects. This requires an adaptation of the assumed contact sensing to make it more similar to the hardware. We expect this to require significant pre-training in simulation and pose a challenging transfer problem from simulation to reality.

\bibliographystyle{plainnat}
\bibliography{references}

\end{document}